\documentclass{article}
\usepackage{ijcai19}

\usepackage{times}
\usepackage{soul}
\usepackage{url}
\usepackage[hidelinks]{hyperref}
\usepackage[utf8]{inputenc}
\usepackage[small]{caption}
\usepackage{graphicx}
\usepackage{amsmath, amsthm, amscd, amssymb, amsfonts}
\usepackage{environ} 
\usepackage{mathtools}

\usepackage{booktabs}
\usepackage{algorithm}
\usepackage{algorithmic}
\usepackage{color}
\usepackage{adjustbox}
\usepackage{multirow}
\usepackage{tikz}
\usepackage{tikz}

\newcommand{\ctikz}[1]{$\vcenter{\hbox{#1}}$}

\title{EmotionX-KU: BERT-Max based Contextual Emotion Classifier}

\author{
	Kisu Yang$^1$\and
	Dongyub Lee$^2$\and
	Taesun Whang$^1$\and
	Seolhwa Lee$^1$\and
	Heuiseok Lim$^1$
	\affiliations
	$^1$Korea University, Republic of Korea, $^2$Kakao Corp.\\
	\emails
	$^1$\{willow4, hts920928, whiteldark, limhseok\}@korea.ac.kr,
	$^2$jude.lee@kakaocorp.com
}

\begin{document}
	\maketitle
	
	\begin{abstract}
		We propose a contextual emotion classifier based on a transferable language model and dynamic max pooling, which predicts the emotion of each utterance in a dialogue. A representative emotion analysis task, EmotionX, requires to consider contextual information from colloquial dialogues and to deal with a class imbalance problem. To alleviate these problems, our model leverages the self-attention based transferable language model and the weighted cross entropy loss. Furthermore, we apply post-training and fine-tuning mechanisms to enhance the domain adaptability of our model and utilize several machine learning techniques to improve its performance. We conduct experiments on two emotion-labeled datasets named Friends and EmotionPush. As a result, our model outperforms the previous state-of-the-art model and also shows competitive performance in the EmotionX 2019 challenge. The code will be available in the Github page.\footnote{\begin{footnotesize}\url{https://github.com/KisuYang/EmotionX-KU}\end{footnotesize}}
	\end{abstract}
	
	\section{Introduction}
	% https://arxiv.org/pdf/1612.01556.pdf
	% http://people.cs.pitt.edu/~wiebe/pubs/papers/emnlp05polarity.pdf
	Sentiment analysis, considered as one of the most important methods to analyze real-world communication \cite{picard2000affective}, is a kind of classification tasks to extract subjective information from language.  The traditional sentiment analysis method, \cite{yu2003towards}, returns opinion polarity towards something, but the approach is confined to analyzing just a single sentence or document, regardless of its surrounding information.
	To resolve this issue, \cite{wilson2005recognizing} performed sentiment analysis with datasets that context information has to be considered. Another advanced dataset, Twitter corpus \cite{pak2010twitter}, is built from social media more similar to real-world communication.

	Under this background, \cite{chen2018emotionlines} released an emotion-labeled corpus of multi-party conversations, EmotionLines, for contextual sentiment analysis. One example of the data set is described in Table \ref{example-table}. The example is a situation that a woman and a man are arguing since she has feelings for him but he is watching someone. The last utterance told by Rachel is supposed to be labeled as \textit{Joy} if it is a single sentence, but it has to be labeled as \textit{Anger} considering the whole dialogue context.

\begin{table}[t]
		\centering
		\begin{adjustbox}{width=0.48\textwidth}
			{\renewcommand{\arraystretch}{5}%}
				\begin{tabular}{cllc}
					\specialrule{.1em}{.1em}{.1em}
					%						\hline
					\multicolumn{1}{c}{\multirow{1}{*}{{{\LARGE{\textbf{Turn}}}}}}
					& \multicolumn{1}{c}{\multirow{1}{*}{{\LARGE{\textbf{Speaker}}}}} & \multicolumn{1}{c}{\multirow{1}{*}{{\LARGE{\textbf{Utterance}}}}}& 
					\multicolumn{1}{c}{\multirow{1}{*}{{\LARGE{\textbf{Emotion}}}}}
					\\ \hline
					{\LARGE{1    }}
					& {\LARGE{Ross}} &
					\ctikz{
						\begin{tikzpicture}
						\node[align=left, text height=1ex]{ 
							
							\\{\LARGE{You had no right to tell me you ever had feelings for me.}}};
						\end{tikzpicture} 
					}
					
					&
					
					{\LARGE{Anger}}          
					
					\\ \cline{1-4} 
					{\LARGE{2}}
					& {\LARGE{Ross}} &  \hspace{0.2cm} {\LARGE{I was doing great, with Julie before i found out about you.}} & {\LARGE{Anger}}
					\\ \hline
					{\LARGE{3}}   
					& {\LARGE{Rachel}}  & 
					\ctikz{
						\begin{tikzpicture}
						\node[align=left, text height=2ex]{ 
							%\hspace{0.1cm} 
							\\{\LARGE{Hey, I was doing great before i found out about you. }} 
							\\[5pt]{\LARGE{You think it's easy for me to see you with Julie?}}};
						\end{tikzpicture} 
					}
					& {\LARGE{Anger}}
					\\ \cline{1-4} 
					{\LARGE{4}}
					& {\LARGE{Ross}}  & 
					
					\ctikz{
						\begin{tikzpicture}
						\node[align=left, text height=4ex]{ 
							%\hspace{0.1cm} 
							\\{\LARGE{It, it's too late, I'm with somebody else, I'm happy.  }} 
							\\[5pt]{\LARGE{This ship has sailed.}}};
						\end{tikzpicture} 
					}
					
					& {\LARGE{Sadness}}        
					\\ \hline
					
					{\LARGE{5}}
					& {\LARGE{Rachel}} & 
					\ctikz{            
						\begin{tikzpicture}
						\node[align=left, text height=4ex]{ 
							{\LARGE{Alright, fine, you go ahead and you do that, alright Ross.}} 
							\\[5pt] {\LARGE{Cause I don't need your stupid ship. }}};
						\end{tikzpicture}     
					}
					& {\LARGE{Anger}}
					
					\\
					\cline{1-4}
					{\LARGE{6}}
					& {\LARGE{Ross}} &   \hspace{0.2cm} {\LARGE{Good.       }} & 
					{\LARGE{Anger    }}      
					\\ \specialrule{.1em}{.1em}{.1em}
				\end{tabular}
			}
		\end{adjustbox}
		\caption{\label{example-table} An example of emotion detection in Friends dataset}
	\end{table}	

	%%%%%%%%%%%%%%%%%%%%%%%%%%%%%%%%%%%%%%%%
	\indent During the recent decade, various neural network models have been proposed to perform this task, arising from their promising performance in many classification tasks. Despite its progress, there are two critical problems of learning long-term dependency \cite{bengio1994learning} and processing in parallel \cite{eckmiller1990parallel}.
	
	Recently, \cite{vaswani2017attention} propose a self-attention mechanism that enables to capture long-term dependency and to compute in parallel. The multi-layered self-attention based language model \cite{devlin2018bert} drastically advances performances of several natural language processing tasks. We will discuss the pre-trained language models in detail in Section \ref{sec:related_work}.

	%We apply the powerful language model to the EmotionX shared task described in Section 3. 
	In this paper, we propose the contextual emotion classifier applied to the EmotionX shared task. The proposed model leverages transferable language model and dynamic max pooling to effectively consider each utterance and its context information together. We consider the contextual emotion classification task as a sequence labeling problem in that each utterance has a context in a dialogue. In addition, we propose several machine learning techniques to handle inherent problem of the shared task, which lead to performance gains as a result.
	%{\color{red}The transferable language model and dynamic max pooling based contextual emotion classifier, considering each utterance and its context information together, is explained in Section 4.}
	
	%In addition, we introduce several machine learning techniques to handle inherent problem of the shared task and to enhance its performance in Section 4 and 5.
    %%%%%%%%%%%%%%%%%%%%%%%%%%%%%%%%%%%%%%%%
	
	The contributions of our paper are as follows:
	\begin{itemize}
		\item We suggest a successful method to deal with the intended problems, context understanding, domain adaptation and class imbalance in the EmotionX shared task.
		\item Our model advances the previous state-of-the-art model, EmotionX-AR \cite{khosla2018emotionx}, by replacing the encoder to a self-attention based transferable language model.
	\end{itemize}
	
	This paper is organized as follows. Section \ref{sec:related_work} provides a literature review. Section \ref{sec:model_description} shows the model description. Section \ref{sec:empirical_study} provides the experimental results and analysis. The conclusion is then provided in Section \ref{sec:conclusion}.
	
	%%%%%%%%%%%%%%%%%%%%%%%%%%%%%%%%%%%%%%%%

	\begin{figure*}[t]
		\centering
		\includegraphics[width=0.75\textwidth]{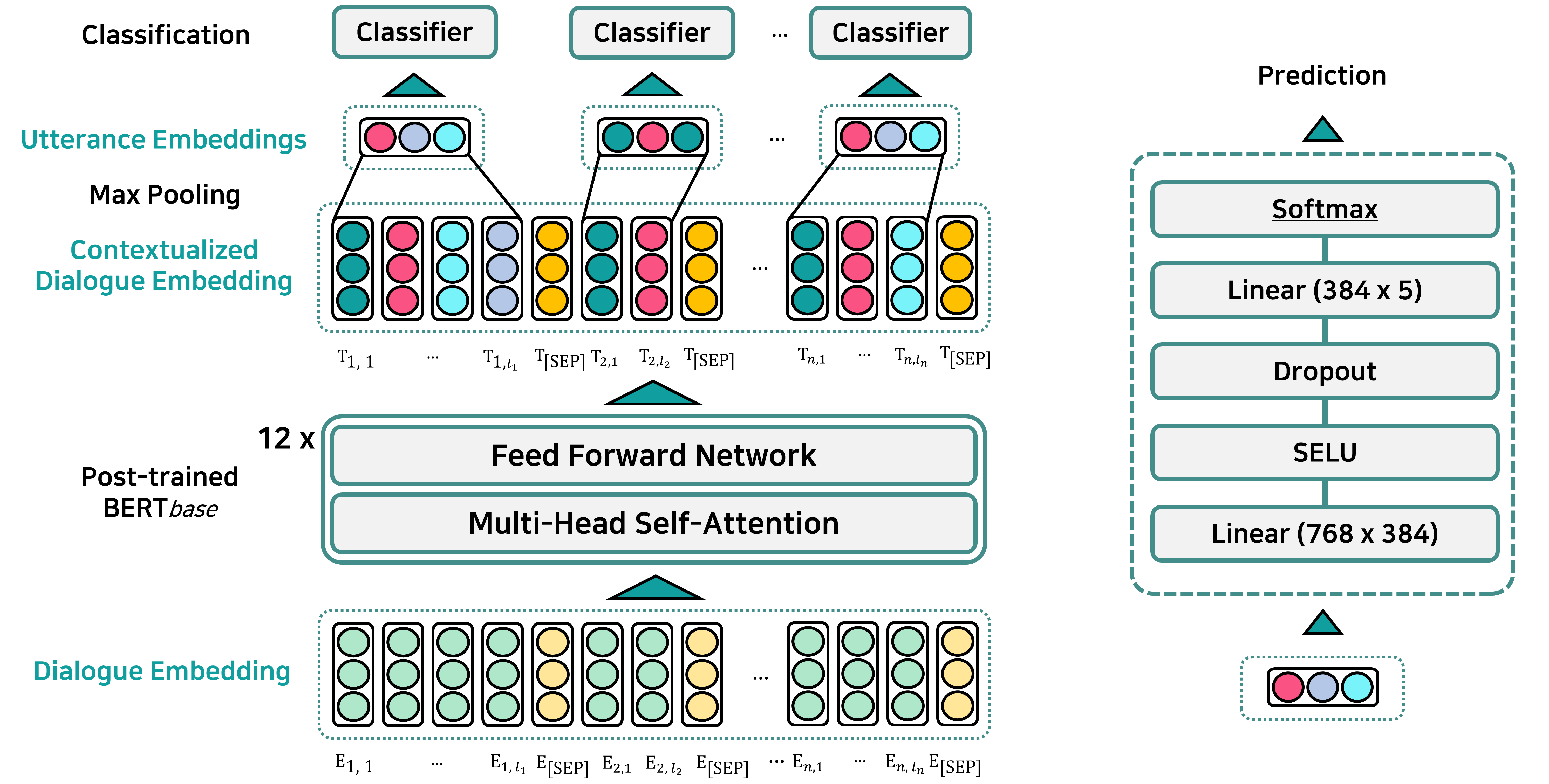}
		\caption{Model overview}
		\label{fig:model_overview}
	\end{figure*}
	
	\section{Related work}
	\label{sec:related_work}
	Our study can be positioned in the connection of the following two topics.
	\subsection{Pre-trained language models}
	Pre-trained langauge models, such as ELMo \cite{peters2018deep}, OpenAI GPT \cite{radford2018improving}, and BERT \cite{devlin2018bert}, have been broadly applied in a variety of NLP tasks  (e.g., sentiment analysis, machine reading comprehension, and textual entailment) and have achieved a great success.
	%Recently, some NLP problems (e.g., question and answering, semantic analysis, reading comprehension, textual entailment) have achieved high performance through pre-trained langauge models such as ELMo \cite{peters2018deep}, OpenAI GPT \cite{radford2018improving} and BERT \cite{devlin2018bert}. 
	They can generate the deep contextualized embeddings since they are pre-trained on a massive unlabeled corpus (i.e., English Wikipedia), thus the method built on top of them can achieve better performance. \\
	\indent Several researches to leverage pre-trained language models on the task of sentiment analysis have been recently proposed. \cite{sun2019utilizing} fine-tuned the BERT model and obtained outperforming results on sentiment analysis task. However, they only focused on single sentence classification and did not conduct an experiment on a conversational corpus where each utterance is semantically related. 
	
	%\textcolor{red}{Unlike what was previously used the pre-trained language model on sentiment analysis},
	One of the previous baselines of emotion detection in a dialogue proposes CNN-DCNN auto-encoder based emotion classifier, but they don't utilize the dialogue context information for predicting emotion of utterance \cite{khosla2018emotionx}.
	Our proposed approach takes \cite{sun2019utilizing}'s and \cite{khosla2018emotionx}'s work one step further by dialogue level with bi-directional pre-trained language model (BERT) on emotion detection, thus the context from both directions on dialogue is reflected much stronger.
	
	\subsection{Sequence labeling}
	
	%{\color{red}
	%The proposed approach is close
	%to (Liu et al., 2015), where only the annotated
	%data for aspect extraction is used. However, we
	%will show that our approach is more effective even
	%compared with baselines using additional supervisions and/or resources.}
	
	Emotion detection task on a multi-party dialogue is similar to dialogue act sequence labeling task. Dialogue acts are semantic labels attached to utterances in a dialogue that present to briefly identify speaker's intention in producing those utterances.
	
	Dialogue acts identification can be interpreted as a sequence labeling problem and can be resolved naively by assigning a label to each element of the sequence independently. Inspired by the sequence labeling problem in a dialogue-level, we apply this concept to our task. For more details, the dialogue-level tokens are input to the post training language model, which takes into account not only their dialogue emotion, thus modeling the dependency among both, labels and utterances, an important consideration of natural dialogue \cite{kumar2018dialogue}.
	
	%(sequence labeling with sentiment)
	More related on our task, \cite{xu2018double} study double embeddings and CNN based sequence labeling for aspect extraction. Aspect extraction is one of the sentiment analysis tasks and aims to extract opinion aspects from opinion based text \cite{yang2018multi}. They achieve very good results on review sentences, but the purpose of these tasks is different from our task in terms of dialogue level. Furthermore, they use pre-trained general-purpose embedding for aspect extraction (e.g., GloVe \cite{pennington2014glove}) while we fine-tune the pre-trained language model to be adapted for the dialogue situation.
	
	To the best of our knowledge, this study is the first report such sequence labeling skeleton, especially on the dialogue level, based pre-trained language model for emotion detection.
	
	\section{Model Description}
	\label{sec:model_description}
	The specification of our model is described in this section.
	
	\subsection{Overview}
	We propose a contextual emotion classifier with a combination of transferable language model and dynamic max pooling as in Figure \ref{fig:model_overview}. (1) Firstly, the input utterances are tokenized according to a byte pair encoding (BPE) algorithm. (2) Then the language model embeds the tokenized inputs into the deep contextualized token representations, which can be sequentially converted to the utterance representations via dynamic max pooling. (3) Finally, the classifier detects contextual emotion, considering the representations.
	%%%%%%%%%%%%%%%%%%%%%%%%%%%%%%%%%%%%%%%%

	\subsection{Input Embedding}
	Every utterance is lower-cased and tokenized by the BPE tokenizer, and all the tokens in the same dialogue are appended with inserting the special token, {\tt[SEP]}, between utterances. Exceptively, if we try to append the tokens of the next utterance for building the list of tokens in the dialogue, but it exceeds the preset maximum length of input tokens, the tokens of the utterance are excluded from the list.
	
	The tokens are embedded through WordPiece embeddings \cite{wu2016google}. The input embeddings, e.g., $E_{i,l_{i}}$ where $i$ is the index of utterance and $l_{i}$ is the length of $i$-th utterance, are the summation of the token embeddings and the positional embeddings.
	%obtained by the positional encoding formula (1) and (2).
	
	%\begin{equation}
	%\resizebox{.70\linewidth}{!}{$
	%	\displaystyle
	%	PE_{(pos,2i)}=sin(pos/10000^{2i/d_{model}})
	%	$}
	%\end{equation}
	%\begin{equation}
	%\resizebox{.75\linewidth}{!}{$
	%	\displaystyle
	%	PE_{(pos,2i+1)}=cos(pos/10000^{2i/d_{model}})
	%	$}
	%\end{equation}
	%where $pos$ denotes the position and $i$ denotes the dimension.
	
	\subsection{Language Model Encoder}
	We adopt transformer \cite{vaswani2017attention} based pre-trained language model (BERT) because it alleviates the long-term dependecy problem, helping capture the worthy context information effectively. Also, the language model shows the promising high performance, for the pre-trained parameters are transferable into other tasks.
	
	To enhance the adaptability of the language model, we post-train \cite{xu2019bert} the model via masked language model (MLM) and next sentence prediction (NSP). Then all layers of the model are fine-tuned while training for the mentioned emotion classification task.
	
	The language model converts the tokenized inputs, e.g., $x = (E_{1,1}, E_{1,2}, ..., E_{1,l_{1}})$, into deep contextualized token representations, e.g., $x' = (T_{1,1}, T_{1,2}, ..., T_{1,l_{1}})$. However, each dialogue has different number of utterances, and each utterance has different number of tokens. To handle this problem, we apply a dynamic max pooling technique to create uniform-sized utterance representations from the different number of tokens as in Figure \ref{fig:model_overview}. It is also considered that the max pooling can assist in keeping important information in each dimension.
	
	\subsection{Imbalanced Emotion Classification}
	The utterance representations from the encoder pass through the classifier what consists of two linear layers, one activation function and dropout as in Figure \ref{fig:model_overview}. The activation function, scaled exponetial linear units (SELUs) \cite{klambauer2017self}, includes normalization processing so that gradient descents can converge more quickly. Dropout is applied to prevent overfitting.
	
	The Friends and EmotionPush datasets suffers from a severe class imbalance problem as in Table 2. To deal with it, we use weighted cross entropy (WCE) as a training loss to weight the samples of minority classes as below.
	
	\begin{equation}
	\resizebox{.50\linewidth}{!}{$
		\displaystyle
		L_{WCE}=-\dfrac{1}{N}\sum^{N}_{i=1}{w_{i}\cdot {L_{CE}}_{i}}
		$}
	\end{equation}
	\begin{equation}
	\resizebox{.30\linewidth}{!}{$
		\displaystyle
		w_{i}=\dfrac{\sum^{N}_{i=1}{x_{i}}}{x_{i}}
		$}
	\end{equation}
	\begin{equation}
	\resizebox{.84\linewidth}{!}{$
		\displaystyle
		{L_{CE}}_{i}=-[y_{i}log(p_{i}) + (1-y_{i})log(1-p_{i})]
		$}
	\end{equation}
	where N is the number of classes. $x_{n}$ is the number of samples of class $n$ in a training set. $y_{n}$ is a ground-truth label, and $p_{n}$ is a probability for corresponding class $n$.
	
	%%%%%%%%%%%%%%%%%%%%%%%%%%%%%%%%%%%%%%%%
	
	%%%%%%%%%%%%%%%%%%%%%%%%%%%%%%%%%%%%%%%%
	
	\label{sec:dataset}
	\begin{table*}[t]
	\centering
	\begin{adjustbox}{width=0.95\textwidth, center}
		\centering
		\renewcommand{\arraystretch}{1.2}%
		\begin{scriptsize}
			\begin{tabular}{cc|cccccccc}
				\hline
				
				\multicolumn{2}{c|}{\multirow{2}{*}{\textbf{Dataset}}}  & \multirow{2}{*}{\begin{tabular}[c]{@{}c@{}}\#Dialogues /\\ \#Utterances\end{tabular}}   & \multirow{2}{*}{\begin{tabular}[c]{@{}c@{}}\#Avg. utterances \\per dialogue\end{tabular}}& \multirow{2}{*}{\begin{tabular}[c]{@{}c@{}}\#Avg. length \\ of dialogues\end{tabular}} &
				\multirow{2}{*}{Neutral} &
				\multirow{2}{*}{Joy} &
				\multirow{2}{*}{Sadness} &
				\multirow{2}{*}{Anger} &
				\multirow{2}{*}{Out-Of-Domain} \\
				&&&&&&&&\\
				\hline
				\multirow{2}{*}{Friends} &Training  & 4,000 / 58,012 &  14.50  & 160.92& 45.0\%& 11.8\%& 3.4\% & 5.2\% &34.6\%\\ 
				&Test & 240 / 3296 & 13.73 & 156.38 & 31.4\%&15.3\%& 3.7\%& 4.3\%& 45.3\% \\
				\hline
				\multirow{2}{*}{EmotionPush} &Training & 4,000 / 58,968  & 14.74 & 114.96 & 66.8\% & 14.2\% & 3.5\% & 0.9\% & 14.6\%\\
				&Test & 240 / 3536 & 14.73 & 92.43 &60.7\% & 17.0\% & 3.1\% & 0.8\% & 18.4\%\\
				\hline
				
			\end{tabular}
		\end{scriptsize}
	\end{adjustbox}
	\caption{Corpus statistics and label distributions of Friends and EmotionPush datasets.}
	\label{table:Dataset}
    \end{table*}

	\section{Empirical Study}
	\label{sec:empirical_study}
	We discuss experiment results of our model in this section.
	
	\subsection{Dataset}
	EmotionX 2019 Challenge is a shared task of Social NLP 2019\footnote{\begin{footnotesize}\url{https://sites.google.com/site/socialnlp2019/}\end{footnotesize}} that detects emotion in dialogue utterances. Two datasets\footnote{\begin{footnotesize}\url{https://sites.google.com/view/emotionx2019/shared-task/datasets}\end{footnotesize}} are released for the challenge, and participants are asked to detect the emotion among four labels ({\em{i.e.,}} Neutral, Joy, Sadness, and Anger). One is the Friends dataset \cite{chen2018emotionlines} , which is multi-party conversations collected from one of the famous TV series, and the other is the EmotionPush \cite{Huang2018EmotionPushEA} that contains messages collected from social network messengers({\em{e.g.,}} Facebook).
	\indent Each dataset contains 4,000 dialogues including 1,000 English-language original version and 3,000 augmented versions, which is back-translated by French, German, and Italian, respectively. Each dialogue is composed of several utterances and they are labeled with 7 emotions ({\em{e.g.,}} Neutral, Joy, Sadness, and etc.). Among the emotions, we regard four emotions ({\em{i.e.,}} Fear, Surprise, Disgust, and Non-Neutral) as Out-Of-Domain, since they are not tested on the evaluation phase. \\
	\indent In Table \ref{table:Dataset}, we describe data and label distribution of Friends and EmotionPush datasets. In terms of label distribution for both datasets, \textit{Neutral} are the most common class, followed by \textit{Joy}, \textit{Sadness}, and \textit{Anger}. Both datasets have imbalanced class distribution, and especially the ratio of \textit{Sadness} and \textit{Anger} is very small. For instance, they account for only 3.4\% and 5.2\%, respectively in the Friends dataset. In the case of EmotionPush, \textit{Anger} label accounts for less than 1\% of the training set. \\
	\indent For the evaluation phase, about 3,000 utterances from 240 dialogues are given to predict one of four emotions. The distribution of two data and classes is similar to the training set of each data.

	\subsection{Experimental Setup}
	For training our model, the number of epochs and the batch size are set to 10 and 1, respectively. We shuffle the training set for every epoch and also apply gradient clipping method.
	
	% https://arxiv.org/pdf/1608.03983.pdf
	The learning rate decreases from $\eta^{i}_{max}$ to $\eta^{i}_{min}$ according to a cosine annealing schedule \cite{loshchilov2016sgdr} as follows:
	\begin{equation}
	\resizebox{.91\linewidth}{!}{$
		\displaystyle
		\eta_{t}=\eta^{i}_{min}+\dfrac{1}{2}(\eta^{i}_{max}-\eta^{i}_{min})(1+cos(\dfrac{T_{cur}}{T_{i}}\pi))
		$}
	\end{equation}
	where $i$ denotes the index of the run, and $T_{cur}$ refers to the number of epochs after the last restart. We set $\eta^{i}_{max}$ = $2e^{5}$ as an initial learning rate and adopt the Adam optimizer \cite{kingma2014adam} working with the scheduled learning rate.
	
	We adopt the pre-trained {\tt uncased BERT-Base} model as the transferable language model where maximum input length is 512. The number of combination layers of a multi-head attention and a feed forward neural network, N in Figure \ref{fig:model_overview}, is 12. The language model is post-trained via a next sentence prediction (NSP) task and masked language model (MLM) with released Friends, EmotionPush and Emory\footnote{\begin{footnotesize}\url{https://github.com/emorynlp/emotion-detection}\end{footnotesize}} \cite{zahiri2018emotion} datasets where the number of  is 100,000 steps. The dimension of hidden representations is set to 768, and the internal hidden size of a classification layer is set to 384. The number of classes is five, including four classes and an out-of-domain class.
	
	\subsection{Evaluation Metric}
	To evaluate the performance of prediction, we mainly use micro f1 score equivalent to weighted accuracy (WA) if every data is tagged with only one class like EmotionLines dataset, obtained by the formula below.
	
	\begin{equation}
	\resizebox{.90\linewidth}{!}{$
		\displaystyle
		micro f1 score = \dfrac{micro-precision\cdot micro-recall}{micro-precision + micro-recall}
		$}
	\end{equation}
	
	\iffalse
	\begin{equation}
	\resizebox{.90\linewidth}{!}{$
		\displaystyle
		macro f1 score = \dfrac{macro-precision\cdot macro-recall}{macro-precision +macro-recall}
		$}
	\end{equation}
	\fi
	
	where
	\begin{equation}
	\resizebox{.70\linewidth}{!}{$
		\displaystyle
		micro-precision = \dfrac{\sum\limits_{c\in C}{TP_{c}}}
		{\sum\limits_{c\in C}{(TP_{c}+FP_{c})}}
		$}
	\end{equation}
	\begin{equation}
	\resizebox{.66\linewidth}{!}{$
		\displaystyle
		micro-recall = \dfrac{\sum\limits_{c\in C}{TP_{c}}}
		{\sum\limits_{c\in C}{(TP_{c}+FN_{c})}}
		$}
	\end{equation}
	
	\iffalse
	\begin{equation}
	\resizebox{.73\linewidth}{!}{$
		\displaystyle
		macro-precision = \dfrac{\sum\limits_{c\inC}{(\dfrac{TP_{c}}{TP_{c}+FP_{c}})}}{N_{C}}
		$}
	\end{equation}
	\begin{equation}
	\resizebox{.68\linewidth}{!}{$
		\displaystyle
		macro-recall = \dfrac{\sum\limits_{c\inC}{(\dfrac{TP_{c}}{TP_{c}+FN_{c}})}}{N_{C}}
		$}
	\end{equation}
	\fi
	
	%%%%%%%%%%%%%%%%%%%%%%%%%%%%%%%%%%%%%%%%%%%%%%%%%%%%%%%%%%%%%%%%%%%%%%%%%%%%%%%%
	\subsection{Results and Analysis}
	\begin{table}[t]
		\begin{adjustbox}{width=0.5\textwidth, center}
			\centering
			\renewcommand{\arraystretch}{1.1}%
			\begin{scriptsize}
				\begin{tabular}{cc|cccccc}
					\toprule
					Dataset & Model   & Micro-f1  & Neutral   & Joy  & Sadness & Anger \\
					\midrule
					\multirow{4}{*}{Friends}
					& Base + Mean 	& 77.5      & 85.3      & 72.3 & 50.0    & 53.5  \\
					& Base + Max 	& 77.1      & 85.0      & 71.5 & 49.7    & 59.9  \\ 
					& Post + Mean 	& \textbf{78.4}      & 85.3      & \textbf{73.3} & \textbf{58.1}    & \textbf{61.4}  \\ 
					& Post + Max 	& 77.5      & \textbf{85.5}      & 71.8 & 49.5    & 57.3  \\ 
					\midrule
					\multirow{4}{*}{EmotionPush}
					& Base + Mean 	& 83.7  & 90.4 		& 71.3 & 59.0    & 18.9  \\ 
					& Base + Max 	& 85.0  & 90.6 		& \textbf{73.8} & 61.1    & 29.8  \\ 
					& Post + Mean 	& 84.1  & 90.5 		& 71.3 & 61.5    & 20.0  \\ 
					& Post + Max 	& \textbf{85.6}  & \textbf{91.1}	 	& 73.5 & \textbf{63.4}    & \textbf{30.6}  \\
					\bottomrule
				\end{tabular}
			\end{scriptsize}
		\end{adjustbox}
		\caption{F1-scores for each emotion of our model trained Friends and EmotionPush data seperately, tested on individual test sets}
		\label{table:each_aug_gold}
	\end{table}
	
	\begin{table}[t]
		\begin{adjustbox}{width=0.5\textwidth, center}
			\centering
			\renewcommand{\arraystretch}{1.1}%
			\begin{scriptsize}
				\begin{tabular}{cc|cccccc}
					\toprule
					Dataset & Model    		& Micro-f1  & Neutral   & Joy  & Sadness & Anger \\
					\midrule
					\multirow{4}{*}{Friends}
					& Base + Mean 	& 74.0  & 83.9 		& 63.6 & 49.6    & 53.8  \\ 
					& Base + Max 	& 76.1      & 85.1      & 65.1 & 51.2    & 57.3  \\ 
					& Post + Mean 	& 76.2   & \textbf{85.4}      & 67.5 & \textbf{54.7}    & 55.9  \\ 
					& Post + Max 	& \textbf{77.5}  & \textbf{85.4} 		& \textbf{70.9} & 52.0    & \textbf{59.7}  \\ 
					\midrule
					\multirow{4}{*}{EmotionPush}
					& Base + Mean 	& 84.4  & 90.6 		& 71.5 & 52.9    & 33.3  \\ 
					& Base + Max 	& 86.0  & \textbf{91.6} 		& 73.5 & \textbf{61.9}    & 29.9  \\ 
					& Post + Mean 	& 85.8 	 	& 91.1 		& 71.7 & 60.1    & 24.6  \\ 
					& Post + Max 	& \textbf{86.3} 	& 91.5	 	& \textbf{74.7} & 61.0    & \textbf{36.2}  \\ 
					\bottomrule
				\end{tabular}
			\end{scriptsize}
		\end{adjustbox}
		\caption{F1-scores for each emotion of our model trained Friends and EmotionPush data together, tested on individual test sets}
		\label{table:both_aug_gold}
	\end{table}
	
	\iffalse
	\begin{table}[t]
		\begin{adjustbox}{width=0.4\textwidth, center}
			\centering
			\renewcommand{\arraystretch}{1.1}%
			\begin{scriptsize}
				\begin{tabular}{cc|cccc}
					\toprule
					
					Dataset &
					Metrics &
					Neutral &
					Joy &
					Sadness &
					Anger \\
					
					\midrule
					\multirow{3}{*}{Friends} &P & 78.0& 85.8 &79.3 &76.4 \\ 
					& R & 95.9 & 62.0 & 38.0 & 57.4 \\
					& F1 & 86.0 & 72.0& 51.4 & 65.6 \\
					\midrule
					\multirow{3}{*}{EmotionPush} &P&88.3 &85.9&80.9&61.5 \\
					&R&96.6&64.9&50.0&29.6\\
					&F1& 92.2 & 73.9 & 61.8 & 40.0 \\
					\bottomrule
				\end{tabular}
			\end{scriptsize}
		\end{adjustbox}
		\caption{\textcolor{red}{Evaluation Results on Friends and EmotionPush datasets.}}
		\label{table:lanel_results}
	\end{table}
	
	\begin{table}[t]
		\begin{adjustbox}{width=0.3\textwidth, center}
			\centering
			\renewcommand{\arraystretch}{1}%
			\begin{scriptsize}
				\begin{tabular}{c|cc}
					\toprule
					Dataset& micro-f1  &macro-f1  \\
					\midrule
					Friends & 79.5 & 68.7 \\ 
					EmotionPush  & 87.6 & 67.0  \\
					\bottomrule
				\end{tabular}
			\end{scriptsize}
		\end{adjustbox}
		\caption{\textcolor{red}{Both Gold}}
		\label{table:overall_results}
	\end{table}	
	\fi
	Experiments are conducted with the released Friends and EmotionPush datasets augmented via back-translation as a training set and their gold datasets for evaluation as a test set.
	
	We assume that the transferable language model which is post-trained with the specific domain corpus predicts the contextual emotion more accurately, and Table \ref{table:each_aug_gold} and  \ref{table:both_aug_gold} support our hypothesis.
	
	To convert the different length of tokens into the uniform sized representations, we design two converters, dynamic averaging and dynamic max pooling. Even though the former sometimes shows the better performance than the latter as in Table \ref{table:each_aug_gold}, the overall performance of the latter is better in the case of training together. Thus, we build our model with post-trained language model and the dynamic max pooling.
	
	When training each dataset seperately compared to training them together, the overall scores on EmotionPush increase, but the performance on Friends dataset decreases. We guess that the parameters of the transferable language model pre-trained with formal corpus might be somehow destroyed when fine-tuning chat-based dialogues, EmotionPush.
	
	For submission version, we implements the k-fold cross validation ensemble method to utilize all the datasets most efficiently where k is 5. Our ensemble model labels each utterance with the most voted emotion based on the decision of k models which are trained with different training set and validation set.

	%%%%%%%%%%%%%%%%%%%%%%%%%%%%%%%%%%%%%%%%%%%%%%%%%%%%%%%%%%%%%%%%%%%%%%%%%%%%%%%%

	\section{Conclusion}
	\label{sec:conclusion}
	We proposed the contextual emotion classifier which consists of the transferable language model and dynamic max pooling. Our model successfully alleviates the three inherent problems in the EmotionX shared task, which is to capture contextual information, to understand informal text dialogues and to overcome a class imbalance problem. It outperforms the previous state-of-the-art model and shows competitive performance in the challenge. However, our model cannot consider all the utterances when the number of input tokens exceeds the preset maximum length, so we expect that future work overcomes this problem.
	
	\section*{Acknowledgements}
	This research was supported by the MSIT(Ministry of Science and ICT), Korea, under the ITRC(Information Technology Research Center) support program (IITP-2018-0-01405) supervised by the IITP(Institute for Information \& communications Technology Planning \& Evaluation)
	
	\bibliographystyle{named}
	\bibliography{ijcai19}
	
\end{document}